\documentclass{article}
\usepackage{arxiv}
\usepackage[utf8]{inputenc} 
\usepackage[T1]{fontenc}   
\PassOptionsToPackage{hyphens}{url}
\usepackage[pagebackref=true,breaklinks=true,letterpaper=true,colorlinks,bookmarks=false,citecolor=teal]{hyperref}
\usepackage{url}        
\usepackage{booktabs}   
\usepackage{amsfonts}     
\usepackage{nicefrac}  
\usepackage{microtype}   
\usepackage[dvipsnames, svgnames, x11names]{xcolor} 
\usepackage{tcolorbox}
\usepackage{graphicx}
\usepackage{bbding}
\usepackage{subcaption}
\usepackage{wrapfig}
\usepackage{bm}
\usepackage{listings}
\usepackage[ruled,vlined]{algorithm2e}

\usepackage{amsmath}
\usepackage{amssymb}
\usepackage{mathtools}
\usepackage{amsthm}
\usepackage{multirow}
\usepackage[capitalize,noabbrev]{cleveref}
\theoremstyle{plain}

\theoremstyle{definition}

\theoremstyle{remark}

\usepackage{accents}
\usepackage[textsize=tiny]{todonotes}

\usepackage[utf8]{inputenc}
\usepackage[english]{babel}
\usepackage{mathrsfs}
\usepackage{times}
\usepackage{latexsym}
\usepackage{comment}
\usepackage{pifont}
\usepackage{lipsum}

\usepackage[]{mathtools}   
\def\$#1\${\begin{align*}#1\end{align*}}
\usepackage{enumitem}
\usepackage[]{amsthm} 
\usepackage[]{amssymb}
\usepackage{bbm}
\usepackage{xcolor}
\usepackage{colortbl}
\definecolor{best}{HTML}{BAFFCD}
\definecolor{issue}{HTML}{FFC8BA}
\definecolor{bad}{HTML}{FFC87C}

\usepackage{amsmath,amsfonts,bm,amsthm,dsfont}

\def\eqref#1{equation~\ref{#1}}

\def\1{\bm{1}}
\def\0{\bm{0}}

\DeclareMathAlphabet{\mathsfit}{\encodingdefault}{\sfdefault}{m}{sl}
\SetMathAlphabet{\mathsfit}{bold}{\encodingdefault}{\sfdefault}{bx}{n}
\usepackage{tcolorbox}
\definecolor{grey}{rgb}{0.33, 0.33, 0.33}

\newcommand{\squishlist}{
\begin{list}{{{\small{$\bullet$}}}}
{\setlength{\itemsep}{1pt}      \setlength{\parsep}{0pt}
\setlength{\topsep}{-2pt}       \setlength{\partopsep}{0pt}
\setlength{\leftmargin}{1em} \setlength{\labelwidth}{1em}
\setlength{\labelsep}{0.5em} } }
\newcommand{\squishend}{  \end{list}  }

\makeatletter
\renewcommand*\env@matrix[1][*\c@MaxMatrixCols c]{%
  \hskip -\arraycolsep
  \let\@ifnextchar\new@ifnextchar
  \array{#1}}
\makeatother

\def\tr{\mathop{\text{tr}}\kern.2ex}

\long\def\comment#1{}

\def\tr{\mathop{\text{Tr}}}

\newcommand{\bel}{\begin{eqnarray}\label}
\newcommand{\eel}{\end{eqnarray}}
\newcommand{\bes}{\begin{eqnarray*}}
\newcommand{\ees}{\end{eqnarray*}}

\usepackage{xcolor}

\ifodd 1

\newcommand{\clar}[1]{\textbf{\color{green}(NEED CLARIFICATION: #1)}}
\newcommand{\response}[1]{\textbf{\color{magenta}(RESPONSE: #1)}}
\else

\newcommand{\com}[1]{}
\newcommand{\clar}[1]{}
\newcommand{\response}[1]{}
\fi
\newcommand{\RNum}[1]{\uppercase\expandafter{\romannumeral #1\relax}}
\newcommand{\methodname}{TAPTRv2}
\newcommand{\fulltitle}{\methodname: Attention-based Position Update Improves\\Tracking Any Point}
\title{\fulltitle} 

\stepcounter{footnote}  % counter + 1.

\author{
    \textbf{Hongyang Li}$^{1,2}$\qquad
    \textbf{Hao Zhang}$^{2,3}$  \qquad
    \textbf{Shilong Liu}$^{2,4}$  \qquad 
    \textbf{Zhaoyang Zeng}$^{2}$ \\
    \textbf{Feng Li}$^{2,3}$\qquad
    \textbf{Tianhe Ren}$^{2}$\qquad
    \textbf{Bohan Li}$^{5}$\qquad
    \textbf{Lei Zhang}$^{1,2}$\thanks{Corresponding author.} \\
    $^1$South China University of Technology. \\
    $^2$International Digital Economy Academy (IDEA). \\
    $^3$The Hong Kong University of Science and Technology. \\
    $^4$Dept. of CST., BNRist Center, Institute for AI, Tsinghua University. \\
    $^5$Shanghai Jiao Tong University. \\
    \vspace{0.9em}
}

\begin{document}

\footnotetext{This work was done while Hongyang Li, Hao Zhang, Shilong Liu, and Feng Li were interns at IDEA.}

\maketitle

\newcommand{\myPara}[1]{\vspace{.05in}\noindent\textbf{#1}}

\begin{abstract}

In this paper, we present {\methodname}, a Transformer-based approach built upon TAPTR for solving the Tracking Any Point (TAP) task. TAPTR borrows designs from DEtection TRansformer (DETR) and formulates each tracking point as a point query, making it possible to leverage well-studied operations in DETR-like algorithms. TAPTRv2 improves TAPTR by addressing a critical issue regarding its reliance on cost-volume, which contaminates the point query’s content feature and negatively impacts both visibility prediction and cost-volume computation. In {\methodname}, we propose a novel attention-based position update (APU) operation and use key-aware deformable attention to realize. For each query, this operation uses key-aware attention weights to combine their corresponding deformable sampling positions to predict a new query position. This design is based on the observation that local attention is essentially the same as cost-volume, both of which are computed by dot-production between a query and its surrounding features. By introducing this new operation, {\methodname} not only removes the extra burden of cost-volume computation, but also leads to a substantial performance improvement. TAPTRv2 surpasses TAPTR and achieves state-of-the-art performance on many challenging datasets, demonstrating the superiority.

\end{abstract}
\section{Introduction}
Tracking any point (TAP) in videos is a more fine-grained task compared to tracking objects using bounding boxes~\cite{meinhardt2022trackformer,sun2012transtrack,xu2022transcenter,zeng2022motr} or their instance masks~\cite{caelles2017one, pont20172017, xu2018youtube, yang2021associating, voigtlaender2019feelvos, oh2019video}. As point correspondence and its visibility prediction in long video sequence is fundamental to many downstream applications, such as augmented reality, 3D reconstruction, and visual imitation~\cite{vecerik2023robotap}, TAP has received increasing attention in the past few years~\cite{harley2022particle, doersch2023tapir, karaev2023cotracker, li2024taptr}.

Some works solve TAP from the 3D perspective~\cite{luiten2023dynamic, duisterhof2023md, yin20234dgen, gu2023videoswap, kratimenos2023dynmf, xiao2024spatialtracker, wang2023tracking}, where they learn an underlying 3D representation of the scene and enable it to transform over time. 
Although such an approach has obtained impressive results, the learning of the 3D representation is nontrivial and challenging. 
Thus most methods are not general and have to be fine-tuned for each video.

To develop a more general solution while keeping a good performance, some methods~\cite{doersch2023tapir, doersch2022tap, harley2022particle, zheng2023pointodyssey, neoral2024mft, moing2023dense} solve the TAP task in 2D space directly. 
Building upon existing optical flow methods~\cite{teed2020raft, Xu_Ranftl_Koltun_2017, Sun_Yang_Liu_Kautz_2018, Wang_Zhong_Dai_Zhang_Ji_Li_2020, Jiang_Lu_Li_Hartley_2021, Huang_Shi_Zhang_Wang_Cheung_Qin_Dai_Li, Shi_Huang_Li_Zhang_Cheung_See_Qin_Dai_Li_2023, Zhao_Zhao_Zhang_Zhou_Metaxas}, especially RAFT~\cite{teed2020raft}, such methods jointly estimate optical flow and point visibility across multiple frames. Supplemented with temporal processing methods such as sliding windows, they achieve remarkable results.
However, these methods are largely affected by previous optical flow estimation methods and model each tracking point as a concatenation of multiple features, including point flow vector, 
point flow embedding, point visibility, point content feature, and local correlation as cost volume~\cite{li2024taptr}. These features normally have clear physical meanings in optical flow, but are simply concatenated and sent as a blackbox vector to MLPs or Transformers and expect MLPs or Transformers to decipher and utilize the features~\cite{doersch2023tapir, doersch2022tap, harley2022particle, zheng2023pointodyssey, karaev2023cotracker}. Such a black box modeling not only makes the model cluttered, but also hinders its optimization and learning efficiency.

\begin{figure*}[t]
    \vspace{-0.1mm}
    \centering
        \includegraphics[width=1\linewidth]{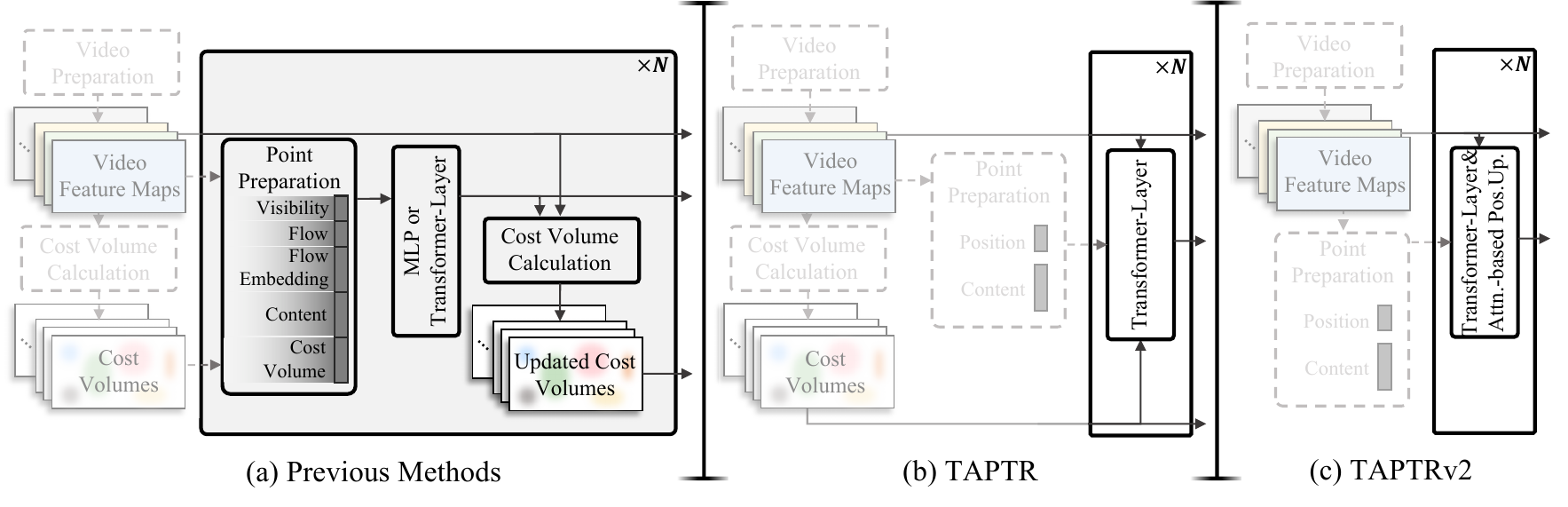}
    \vspace{-6mm}
    \caption{
    Comparison of the frameworks among previous works, TAPTR, and TAPTRv2. Inspired by DETR-based detection algorithms, TAPTR formulates the point tracking problem as a detection problem and simplifies the overall pipeline to a well-studied DETR-like framework. After introducing the attention-based position update operation into Transformer decoder layers, 
    the overall pipeline is further simplified to be as straightforward as detection methods. The operations within dashed boxes are executed only once.
    }
    \vspace{-5mm}
    \label{fig.comparison}
\end{figure*}

To more effectively utilize the features, TAPTR takes inspiration from DEtection TRansformer (DETR)~\cite{carion2020end, meng2021conditional, liu2022dab} and models each tracking point as a point query as in DETR with a content part and a positional part (point coordinates). Each query is refined layer by layer, with its visibility predicted by its updated content feature. Point queries exchange information through spatial or temporal attention in the same frame or along the temporal dimension. 
Such a point query formulation not only makes the TAP pipeline conceptually simple, but also lead to a remarkable performance.

However, despite its demonstrated performance improvement, TAPTR still relies on the cost-volume feature and has a questionable design, which concatenates the cost-volume feature of a point query and its content part, followed with an MLP transformation (See Eq. 4 in ~\cite{li2024taptr}). As after each Transformer decoder layer, the updated point query needs to predict a relative position to update the query's coordinates, aggregating cost-volume, which is a local correlation information, to the query's content part helps the point query predict a more accurate position. However, aggregating cost-volume also contaminates the query's content part, which has two negative impacts. First, the cross-attention operation in each Transformer decoder layer needs to compute attention maps, which are the similarities between point queries and image feature keys\footnote{Note that in TAPTR, deformable attention is used, which can be considered as a sparse and approximate attention as its attention weights are directly predicted based the feature of a query without comparing the query with image features. Here we use dense attention for discussion for its simple and clear definition.}. Yet queries and keys have different formulations. While both queries and keys have their content part and positional part, queries are contaminated by cost-volume whereas keys are not. Such a difference makes the attention computation implausible. Second, a contaminated point query also yields to inaccurate cost-volume as the computation of cost-volume also needs to compare the point query with its local image features. 
The experiments in TAPTR show that, with such contaminated cost-volumes, the performance will suffer a big drop.
Moreover, the incorporation of cost-volume in TAPTR not only results in redundant computations, but also leaves the simplicity one step behind query-based object detection methods~\cite{zhang2022dino, liu2022dab, meng2021conditional, li2022dn}. This raises several intriguing questions: Why is cost-volume necessary? Is there any alternative that can be developed without redundant effort? How can the cost-volume or its alternative be better utilized without contaminating a point query?

With this motivation, we propose {\methodname}. Compared to TAPTR, {\methodname} does not aggregate cost-volume to queries to avoid contaminating their content features. Meanwhile, with a deeper analysis recognizing the importance of the information captured by cost-volume, we propose a novel Attention-based Position Update (APU) operation, which, for each query, uses its local attention weights to combine its local relative positions to predict a new query position. Such an operation is equivalent to a cross-attention operation from a point query (Q) to image features (K) using local attention, but the values are local relative positions (V) instead of image features. This design is based on the observation that local attention is essentially the same as cost-volume, both of which are computed by dot-production between a query and its surrounding features.
% with a difference that local attention is softmax-normalized. 
By introducing this new operation, the TAP framework is further simplified in {\methodname}, which not only removes the extra burden of cost-volume computation, but also yields a substantial performance improvement.

In our implementation, we follow TAPTR and adopt deformable attention for its proven efficiency and effectiveness in DETR-based detection algorithms. However, as deformable attention directly predicts attention weights for a query without comparing the query with image features, we use its variant, key-aware deformable attention~\cite{li2023lite} which computes attention weights by explicitly comparing a query with image features. Our ablation studies show that key-aware deformable attention is indeed more effective as it precisely matches the design of attention-based position update.

As shown in Fig.~\ref{fig.comparison}, with the help of our analysis and our simple yet effective designs, TAPTRv2 is much simpler and clearer than previous methods. To further verify the superiority of TAPTRv2 brought by our clear point query design, we conduct experiments on several TAP datasets, TAPTRv2 achieves the best performance on all of the datasets. 

\section{Related Work}
\noindent\textbf{Optical Flow Estimation.} 
Optical flow is a long-standing problem in computer vision, which has attracted a great amount of research~\cite{Horn_Schunck_1981, Black_Anandan_2002, Bruhn_Weickert_Schnörr_2005} over the past few decades. Particularly, in the last decade, deep learning-based methods~\cite{Dosovitskiy_Fischer_Ilg_Hausser_Hazirbas_Golkov_Smagt_Cremers_Brox_2015, Ilg_Mayer_Saikia_Keuper_Dosovitskiy_Brox_2017, Xu_Ranftl_Koltun_2017, Sun_Yang_Liu_Kautz_2018, Wang_Zhong_Dai_Zhang_Ji_Li_2020, Jiang_Lu_Li_Hartley_2021, Xu_Yang_Cai_Zhang_Tong_2021, Zhang_Woodford_Prisacariu_Torr_2021, Huang_Shi_Zhang_Wang_Cheung_Qin_Dai_Li, Shi_Huang_Li_Zhang_Cheung_See_Qin_Dai_Li_2023, Zhao_Zhao_Zhang_Zhou_Metaxas} have demonstrated a strong advantage in this field. DCFlow~\cite{Xu_Ranftl_Koltun_2017} was the first to verify the feasibility of using cost-volume to address the optical flow problem. The robustness of cost-volume has enabled many subsequent works~\cite{Sun_Yang_Liu_Kautz_2018, Wang_Zhong_Dai_Zhang_Ji_Li_2020, teed2020raft} and dominated this field. However, optical flow estimation methods can only handle flow estimation between two frames, which prevents them from utilizing long-term temporal information to improve accuracy. More importantly, in the presence of occlusions, optical flow methods often suffer from the problem of tracking target change. These issues make it challenging for optical flow estimation methods to process videos directly.

\noindent\textbf{Tracking Any Point.} The TAP task is defined to estimate the flow of any point between any two consecutive frames and predict the visibility of the tracked point in every frame in the entire video. Some works~\cite{wang2023omnimotion,xiao2024spatialtracker,song2024track} aim to address the TAP task by constructing a time-varying 3D field. Due to the difficulty of learning a 4D field, such methods have to retrain their network to fit each video, which is normally too slow and impractical for many applications.
Given the similarities between TAP and optical flow, most current  methods~\cite{harley2022particle,zheng2023pointodyssey,doersch2022tap,doersch2023tapir,karaev2023cotracker} follow the optical flow methods, especially RAFT~\cite{teed2020raft}, but extend to multi-frame scenarios. By contrast, TAPTR~\cite{li2024taptr} takes inspiration from Transformer-based object detection algorithms and models point tracking as a point detection problem, which makes TAP conceptually simple and leads to a remarkable performance improvement.

\section{{\methodname}}

\subsection{Overview}

\begin{figure*}[t]
    \centering
        \includegraphics[width=0.8\linewidth]{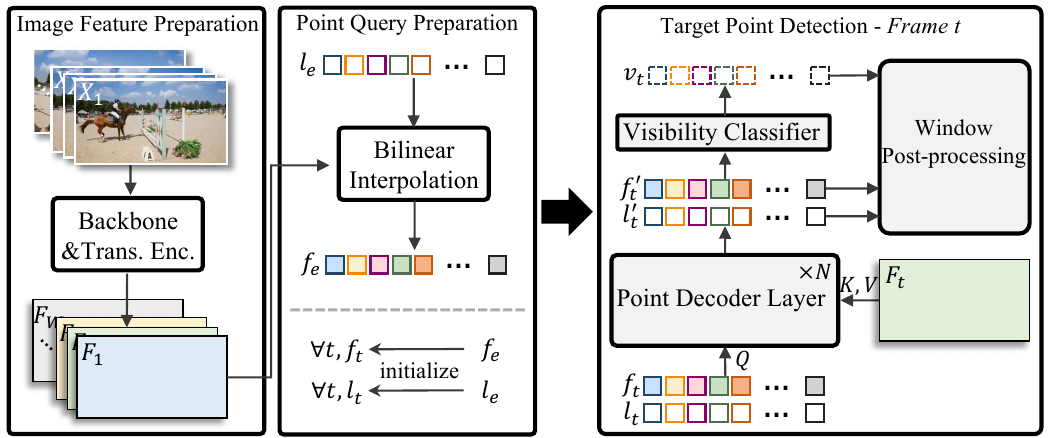}
    \caption{
    The overview of {\methodname}. 
    The image feature preparation part and the point query preparation part prepare the image features of each frame of an input video and the point queries for each tracking point in every frame. 
    The target point detection part takes the prepared image features and point queries as input. For every frame, each point query aims to predict the position and visibility of its target point.
    }
    \vspace{-3mm}
    \label{fig.pipeline}
\end{figure*}

As shown in Fig.~\ref{fig.pipeline}, {\methodname} shares a similar architecture to DETR-based object detection. More specifically, its point query bears a strong resemblance to queries designed for visual prompt-based object detection~\cite{li2023visual, jiang2024t}. Thus {\methodname} mainly consists of three parts, image feature preparation, point query preparation, and target point detection. 
To process videos of dynamic lengths, we follow previous works~\cite{harley2022particle, karaev2023cotracker, doersch2023tapir, doersch2022tap, li2024taptr} and utilize the sliding window strategy, which divides a video into windows of lengths $W$ and processes $W$ frames in parallel once at a time. 
Since {\methodname} is built upon TAPTR, to make this section self-contained, we will first provide a brief overview of the TAPTR framework and then describe how {\methodname} improves TAPTR.

\noindent\textbf{Image Feature Preparation.}
Our method is orthogonal to any vision backbones. In this work, we use ResNet-50 as our backbone as it is the most widely used backbone for fair comparison in DETR-related research works~\cite{zhang2022dino, li2022dn, liu2022dab, meng2021conditional}. After obtaining multi-scale image feature maps from the image backbone, we send them into a Transformer-encoder to further enhance the features as well as the receptive fields of image features.
After that, each frame $X_t$ is ended up with 
a set of high-quality multi-scale image feature maps $F_t$,.

\noindent\textbf{Point Query Preparation.}
Considering general TAP application scenarios, each tracking point has its unique start frame and initial position. We define their initial locations as $l_e = \{l_{e^i}^i\}_{i=1}^N$, where $N$ is the number of points to be tracked, $e^i$ indicates the start frame ID when the $i$-th  tracking point first emerges or starts to be tracked.
Similar to the visual prompt-based detection methods~\cite{li2023visual,jiang2024t}, {\methodname} needs to prepare a visual feature to describe each target tracking point. Following previous methods~\cite{li2024taptr, karaev2023cotracker, harley2022particle, doersch2023tapir}, without loss of generality, for the $i$-th target tracking point, its initial feature $f_e^i$ can be obtained by conducting bilinear interpolation on 
the multi-scale feature maps of its start frame $F_{e^i}$ at its initial position $l_{e^i}^i$. Then the sampled results are transformed using an MLP to fuse multi-scale information.
Since the tracking of a target point across a video can be treated as detecting the target point in every frame of the video. 
Following the formulation of object queries in DETR-based object detection methods, for every video frame, each point query consists of a content part and a positional part, i.e. $Q_t^i = \left(f_t^i, l_t^i\right)$, which are initialized with the prepared initial feature and location of its corresponding target tracking point
\begin{equation}
        \forall 1\leq  i \leq N, \forall 1\leq  t \leq T, Q_t^i = (f^i_{t}, l^i_{t}) \Leftarrow (f^i_{e}, l^i_{e}).
% \vspace{-1mm}
\end{equation}

\noindent\textbf{Target Point Detection in Every Frame.}
After preparing the image features of every frame and every point query in each frame, the TAP task can be clearly formulated as point detection. Taking the $t$-th frame for example, we treat its image features $F_t$ as keys and values, the point queries $(f_t, l_t)$ as queries, and send them to a sequence of Transformer decoder layers. In every Transformer decoder layer, both the content part and positional part of the point queries will be refined. After the multi-layer refinement, the final positional part $l_t^{'}$ of each point query is treated as the predicted position of its corresponding target tracking point in the $t$-th frame. Meanwhile, the content part is used to predict the visibility of the tracking point using an MLP-based visibility classifier

\begin{equation}
    v_t = \texttt{Vis}(f_t^{'}).
\end{equation}

\noindent\textbf{Window Post-Processing.}
After obtaining the detection results of all point queries in a window, each tracking point's trajectory and visibility in this window can be updated. To proceed with the next window, we use the predicted tracking point positions and their corresponding content features in the last frame of the current window to initialize point queries in the next window. This simple strategy effectively propagates the latest prediction result to the next window.

\subsection{Analysis of Cost Volume Aggregation in TAPTR Decoder}
\begin{figure*}[t]
    \vspace{-0.1mm}
    \centering
        \includegraphics[width=1\linewidth]{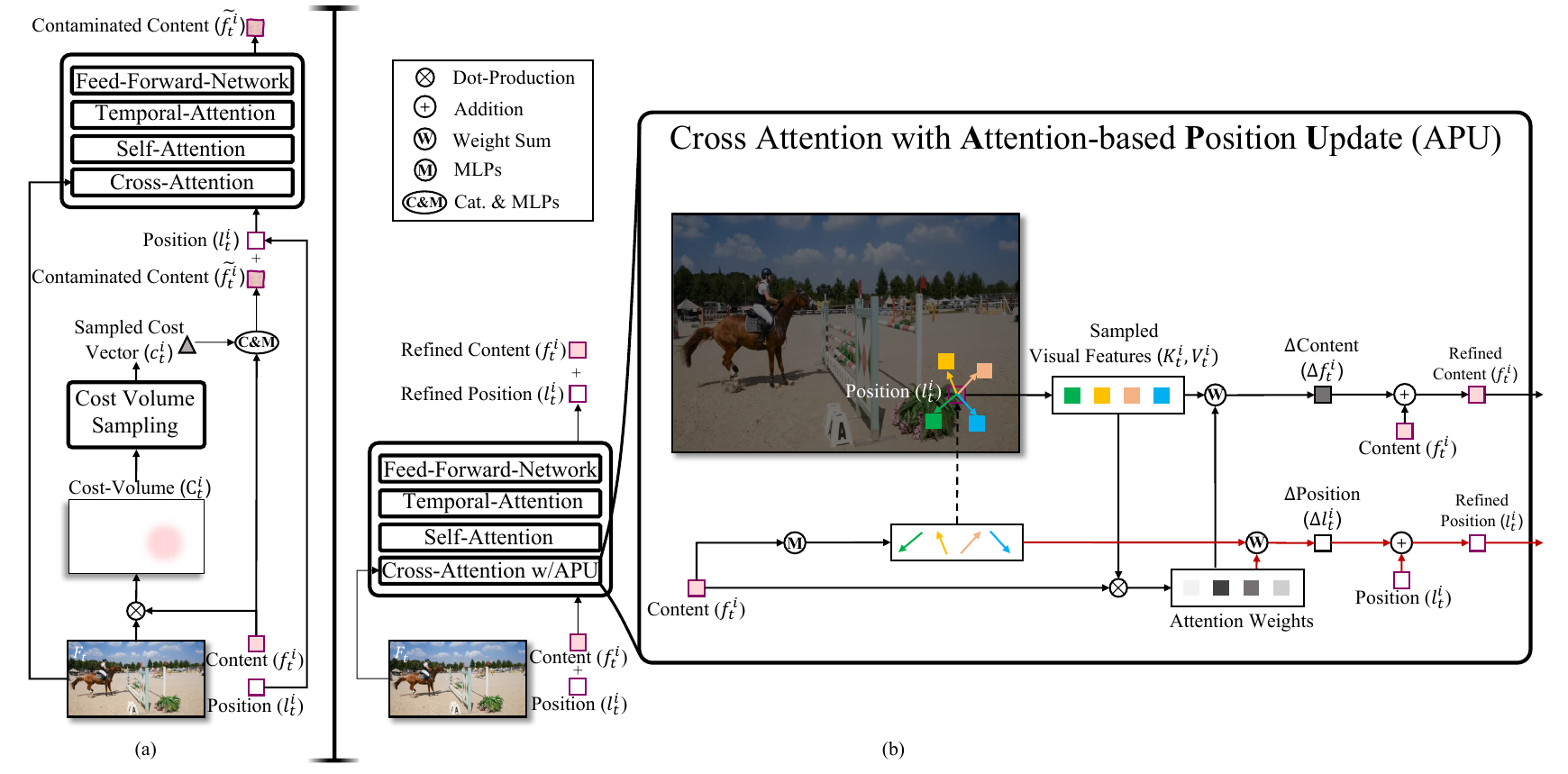}
    \caption{
    Comparison of the decoder layer in TAPTR (a) and TAPTRv2 (b). In TAPTR (a), cost-volume aggregation will contaminate the content feature, affecting cross-attention and leading to the contaminated cost-volume in the next layer. In TAPTRv2 (b), with the introduction of Attention-based Position Update (APU) in cross attention, not only the attention weights are properly used to update the position of each point query and mitigate the domain gap, but also the content feature of each point query is kept uncontaminated, which is crucial for visibility prediction.
    We use an RGB image to represent the multi-scale feature maps for better visualization.
    }
    \label{fig.cross_attention_with_position_update}
\end{figure*}

TAPTR regards cost-volume as indispensable and adds extra cost-volume aggregation blocks before sending point queries to Transformer decoder layers. The extra block for cost-volume not only contaminates the point queries' content feature but also makes the pipeline complex as in Fig~\ref{fig.cross_attention_with_position_update} (a).

\noindent\textbf{Cost Volume Aggregation.}
Taking the $i$-th point query $Q_t^i$ in the $t$-th frame as an example,
TAPTR conducts dot-production between $Q_t^i$ and the image feature maps $F_t$ of the $t$-th frame to obtain the point query's cost-volume $C_t^i$. With the help of grid sampling, TAPTR obtains sampled cost vector $c_t^i$ from $C_t^i$ around the location of the point query $l_t^i$. 

\noindent\textbf{Contaminating Content Feature.}
After obtaining $c_t^i$, it is fused into the point query's content feature $f_t^i$ through an $\texttt{MLP}$
\begin{equation}
\label{eq.blurred_content}
    \tilde{f_t^i} \Leftarrow \texttt{MLP}\left(\texttt{Cat}\left(f_t^i, c_t^i\right)\right),
\end{equation}
where \texttt{Cat} denotes concatenation along the channel dimension, $\tilde{f_t^i}$ indicates the contaminated content feature.
Although such a fusion makes use of the cost volume, the point query's content feature, which is expected to describe its target tracking point's visual feature, is contaminated. The contamination will further affect the calculation of cost volume in the next layer, preventing TAPTR from using more accurate cost-volume. The ablation study in TAPTR verifies that, if TAPTR updates the cost volume in every decoder layer, the performance will drop significantly.

\begin{table*}[tb]
\begin{center}
\resizebox{0.9\linewidth}{!}{ %
\begin{tabular}{c|c|c|c|ccc|ccc}
\toprule
    & Self       & Temporal    & Cost     & \multicolumn{3}{c|}{DAVIS (Out of Domain)} & \multicolumn{3}{c}{Kubric (In Domain)} \\
Row & Attention  & Attention   & Volume   &  AJ & $<\delta^{x}_{avg}$ & OA &  AJ & $<\delta^{x}_{avg}$ & OA \\
\midrule

1 &\ding{55}  &\ding{55}    &\ding{55}  &47.4& 62.2& 82.5& 79.7& 87.8& 94.3 \\ 
\midrule
2 &\checkmark &\ding{55}    &\ding{55}  &50.6 ($\uparrow$3.2)& 64.5& 85.7& 83.7 ($\uparrow$4.0)& 90.8& 95.7 \\
3 &\ding{55}  &\checkmark   &\ding{55}  &54.3 ($\uparrow$6.9)& 68.3& 87.0& 83.4 ($\uparrow$2.7)& 90.6& 96.5 \\
4 &\ding{55}  &\ding{55}    &\checkmark &52.0 ($\uparrow$4.6)& 66.3& 84.7& 79.5 ($\downarrow$0.2)& 87.9& 94.6 \\

\bottomrule
\end{tabular}
}
\caption{We start with a baseline (Row 1) without using self-attention, temporal-attention, and cost-volume, and add each component from TAPTR in turn to show their impact on in-domain and out-of-domain datasets. The addition of self-attention and temporal attention leads to a significant improvement on both the in-domain and out-of-domain datasets. However, the addition of cost-volume only leads to a significant improvement on the out-of-domain dataset but a negative impact on the in-domain dataset, showing that the importance of cost-volume mainly comes from its ability to mitigate the domain gap. Note that the in-domain evaluation set is created by rendering additional 150 videos using the same setting as the training set.}
% \vspace{-12mm}
\label{tab:analyze_domain_gap}
\end{center}
\end{table*}

\noindent\textbf{Cost-volume Necessity Analysis.} 
\label{sec.analyze_costvolume}
Although the use of cost-volume leads to a questionable feature contamination problem, cost-volume still contributes to the performance greatly in TAPTR. 
To understand the reason why cost-volume is necessary, we conduct an ablation study on TAPTR. As shown in Table~\ref{tab:analyze_domain_gap}, we remove the self-attention, temporal-attention, and cost-volume components from TAPTR's decoder, and add them one by one and observe their impact on the performance of in-domain and out-of-domain datasets. The results show that both self-attention and temporal-attention bring significant improvement on both in-domain and out-of-domain datasets. However, while cost-volume also brings a significant improvement on the out-of-domain dataset, it leads to a slightly negative effect (0.2 AJ drop) on the in-domain dataset.
This contradictory result indicates that cost-volume is only essential for mitigating the domain gap and enhancing the generalization capability of the model.
This is quite reasonable because cost-volume is essentially the information of similarities between features, which is why it is called correlation map in some works~\cite{karaev2023cotracker, teed2020raft, harley2022particle, zheng2023pointodyssey}. Due to the domain gap, the features learned by a TAP model can hardly be generalized to out-of-domain datasets. In comparison, the correlation information is more robust to domain changes as it captures the similarity information between local features. This motivates us to design a more effective approach to utilizing cost-volume, which we find is equivalent to attention weight in essence.

\subsection{Cross Attention with Attention-based Position Update}
According to our analysis in Sec.~\ref{sec.analyze_costvolume}, the effectiveness of cost-volume comes from its robust deep feature similarity, which is also in essence equivalent to how attention weights are computed. To leverage this insight, we still choose the deformable operation for its computational efficiency in using multi-scale image features, but replace its attention prediction with key-aware attention prediction, which is called key-aware deformable attention~\cite{li2023lite}.

\noindent\textbf{Key-Aware Deformable Attention Revisiting.}
Deformable attention directly predicts the attention weights for a query without comparing the query with image features. While this design is proven effective in object detection, it is inappropriate for TAP, as we want to leverage the attention weights as a replacement of cost-volume. Using key-aware deformable attention meets this need. Taking $Q_t^i$ as an example, key-aware deformable attention can be formulated as
\begin{equation}
\label{eq.key_aware}
    \begin{gathered}
    S_t^i = W^S\cdot f_t^i, K_t^i = V_t^i = \texttt{Bili}(F_t, l_t^i + S_t^i), \\
    Q_t^i = f_t^i, A_t^i = f_t^i \cdot K_t^i, \Delta f_t^i = \texttt{SoftMax}(A_t^i / \sqrt{d})\cdot V_t^i\\
    f_t^{i} \Leftarrow f_t^i + \Delta f_t^i,
    \end{gathered}
\end{equation}
where $S_t^i$ denotes the sampling offsets, $Q_t^i$, $K_t^i$, $V_t^i$ and $A_t^i$ indicate the query, key, value, and attention weights inside the attention mechanism, respectively, $W^S$ is a learnable parameter, $d$ is the number of key channels, $\texttt{Bili}$ indicates the bilinear interpolation, $\Delta f_t^i$ is the update of content feature.
Note that, for notation simplicity, we assume there is only one attention head and $F_t$ has only one scale.

\noindent\textbf{Attention-based Position Update.}
Since the attention weights $A_t^i$ in Eq.~\ref{eq.key_aware} reflect the similarity between the point query $Q_t^i$ and the sampled image features (K), the attention weights and their corresponding sampling offsets imply where the target tracking point is in the current frame. Thus we combine the sampling offsets using the computed attention weights to obtain a position update, and the update will be used to update the location of the point query. This is exactly a (sparse) cross-attention operation, in which the sampling offsets are values (V). Note that to update the content part of the point query, there is another cross-attention operation, in which the sampled image features are values (V). These two cross-attention operations can use the same attention weights. However, we empirically find that the sharing of attention weights for content and position update is detrimental to model optimization. We guess the update of content and position may need different distribution of the attention weights (e.g. more spiked or more smooth). Thus, we introduce an $\texttt{MLP}$ to work as a $\texttt{Disentangler}$ to disentangle the weights required for content and position update. The process can be formulated as
\begin{equation}
    \begin{gathered}
    % \Delta l_t^i = \texttt{SoftMax}\left(\texttt{Disentangler}\left(A_t^i\right) / \sqrt{d}\right) \cdot S_t^i \\
    \Delta l_t^i = \texttt{SoftMax}\left(\texttt{Disentangler}\left(A_t^i / \sqrt{d}\right)\right) \cdot S_t^i, \\
    l_t^{i} \Leftarrow l_t^i + \Delta l_t^i,
    \end{gathered}
\end{equation}
where $\Delta l_t^i$ indicates the position update.
Thanks to the separation of cost-volume from the content feature, the content feature can be kept clean, which leads to more accurate point visibility prediction as evidenced in Table~\ref{tab:benchmark_tapvid}. Meanwhile, our proposed attention-based position update operation deliberately utilizes attention weights as an equivalent form of cost-volume to perform position update, which effectively helps mitigate the domain gap problem.

\section{Experiments}
\label{sec:exp}
\vspace{-2mm}
We conduct extensive experiments on multiple challenging evaluation datasets collected from real world to verify the superiority of TAPTRv2. Detailed ablation studies for our main contribution are also provided to show the effectiveness of each design in modeling.

\begin{table*}[t]
\begin{center}
\resizebox{0.83\linewidth}{!}{ %
\begin{tabular}{l|ccc|ccc|ccc}
\toprule
 & \multicolumn{3}{c|}{DAVIS} & \multicolumn{3}{c|}{DAVIS-S} & \multicolumn{3}{c}{Kinetics} \\
Method & AJ & $<\delta^{x}_{avg}$ & OA &  AJ & $<\delta^{x}_{avg}$ & OA &  AJ & $<\delta^{x}_{avg}$ & OA \\
\midrule
PIPs~\cite{harley2022particle} & -- & -- & -- & 42.0 & 59.4 & 82.1 & 31.7 & 53.7 & 72.9 \\
TAP-Net~\cite{doersch2022tap} & 36.0 & 52.9 & 80.1 & 38.4 & 53.1 & 82.3 & 38.5 & 54.4 & 80.6 \\
Context-PIPs~\cite{bian2023context} & -- & -- & -- & 48.9 & 64.0 & -- & -- & -- & 79.8 \\
MFT~\cite{neoral2024mft} & 47.3 & 66.8 & 77.8 & 56.1 & 70.8 & 86.9 & 39.6 & 60.4 & 72.7 \\
TAPIR~\cite{doersch2023tapir} & 56.2 & 70.0 & 86.5 & 61.3 & 73.6 & 88.8 & \underline{49.6} & 64.2 & 85.0 \\
OmniMotion\cite{wang2023tracking} & 52.7 & 67.5 & 85.3 & -- & -- & -- & -- & -- & --  \\
CoTracker-Single\cite{karaev2023cotracker} & 60.6 & 75.4 & 89.3 & 64.8 & 79.1 & 88.7 & 48.7 & \underline{64.3}& \textbf{86.5} \\
CoTracker2-All\cite{karaev2023cotracker} & 60.7 & 75.7 & 88.1 & -- & -- & -- & -- & -- & -- \\
CoTracker2-Single\cite{karaev2023cotracker} & 62.2 & 75.7 & 89.3 & 65.9 & \textbf{79.4} & 89.9 & -- & -- & -- \\
TAPTR~\cite{li2024taptr} & \underline{63.0} & \textbf{76.1} & \underline{91.1} & \underline{66.3} & \underline{79.2} & \underline{91.0} & 49.0 & \textbf{64.4} & 85.2 \\
\midrule
\textcolor[rgb]{0.753,0.753,0.753}{BootsTAP$^\dag$\cite{doersch2024bootstap}}  & \textcolor[rgb]{0.753,0.753,0.753}{61.4} & \textcolor[rgb]{0.753,0.753,0.753}{74.0} & \textcolor[rgb]{0.753,0.753,0.753}{88.4} & \textcolor[rgb]{0.753,0.753,0.753}{66.4} & \textcolor[rgb]{0.753,0.753,0.753}{78.5} & \textcolor[rgb]{0.753,0.753,0.753}{90.7} & \textcolor[rgb]{0.753,0.753,0.753}{54.7} &  \textcolor[rgb]{0.753,0.753,0.753}{68.5} & \textcolor[rgb]{0.753,0.753,0.753}{86.3} \\
\midrule
Ours (TAPTRv2) & \textbf{63.5} & \underline{75.9} & \textbf{91.4} & \textbf{66.4} & 78.8 & \textbf{91.3} & \textbf{49.7} & \underline{64.2} & \underline{85.7} \\
\bottomrule
\end{tabular}
}
\caption{Comparison of {\methodname} with prior methods. Note that, BootsTAP$^\dag$ is a concurrent work and introduces extra 15M video clips from publicly accessible videos for training.}
\vspace{-8mm}
\label{tab:benchmark_tapvid}
\end{center}
\end{table*}

\vspace{-3mm}
\subsection{Datasets and Evaluation Settings}
\vspace{-3mm}
\noindent\textbf{Datasets.}
Following previous works~\cite{li2024taptr, karaev2023cotracker, harley2022particle, doersch2023tapir} we train TAPTRv2 on the Kubric dataset, which consists of 11,000 synthetic videos generated by Kubric Engine~\cite{greff2021kubric}. In each video of Kurbic, Kubric Engine simulates a set of rigid objects falling down the floor from the air and bouncing. In each video, 2,048 points on the surface of background and moving objects are randomly sampled to generate point trajectories for training. 
% Considering that the tracking of the points on the moving objects is much harder than those on the background, relatively more points are sampled on the surface of moving objects. 
During training, for training efficiency, the resolution of the videos is resized to 512$\times$512, and we randomly select 700-800 trajectories for training from each video. 
We evaluate our method on the challenging TAP-Vid-DAVIS~\cite{pont20172017} and TAP-Vid-Kinetics~\cite{carreira2017quo} datasets. Both datasets are from TAP-Vid~\cite{doersch2022tap} and are collected from real world and annotated by well-trained annotators. TAP-Vid-DAVIS has 30 challenging videos with complex motions and large-scale changes of the objections. TAP-Vid-Kinetics has over 1,000 YouTube videos, and the camera shaking and complex environment make it also a challenging dataset.

\vspace{-1mm}
\noindent\textbf{Evaluation Metrics and Settings.}
For evaluation, we follow the metrics proposed in TAP-Vid~\cite{doersch2022tap}, including Occlusion Accuracy (OA) which describes the accuracy of classifying whether the target tracking points are visible or occluded, $<\delta_{avg}^x$ which reflects the average precision of the predicted tracking points' location at thresholds of 1,2,4,8,16 pixels, and Average Jaccard (AJ) which is a comprehensive metric to measure the overall performance of a point tracker from the perspective of both location and visibility classification. 
Meanwhile, there are two evaluation modes to accommodate online and offline trackers. The ``Strided'' mode is for offline trackers. The ``First'' mode is for online trackers and is much harder. In this paper, without specification, we evaluate our method on the ``First'' mode, and to facilitate comparisons with offline methods, we follow previous methods~\cite{karaev2023cotracker,li2024taptr} to further report our performance on TAP-Vid-DAVIS dataset in the ``Stride'' mode.
Note that, since the resolution of the input image has a great influence on the performance, for fair comparison, we follow previous works to limit the resolution of our input image to 256$\times$256.

\vspace{-3mm}
\subsection{Implementation Detail}
\vspace{-2mm}
We follow the previous work~\cite{li2024taptr} and use ResNet-50 as the image backbone for both experimental efficiency and fair comparison. We employ two Transformer encoder layers with deformable attention~\cite{zhu2020deformable} to enhance feature quality, and five Transformer decoder layers by default to achieve the results that are fully optimized. We use AdamW~\cite{zhuang2022understanding} and EMA~\cite{klinker2011exponential} for training. We use 8 NVIDIA A100 GPUs, accumulating gradients 4 times to approximate a total batch size of 32, and train TAPTRv2 for approximately 44,000 iterations.

\vspace{-3mm}
\subsection{Comparison with the State of The Arts}
\vspace{-2mm}
We compare TAPTRv2 with previous methods on TAP-Vid-DAVIS and TAP-Vid-Kinetics to show its superiority in online tracking. To broaden our comparison, we also present the performance of TAPTRv2 in the ``Strided'' mode on DAVIS dataset (DAVIS-S). The results in Table~\ref{tab:abl_attn_pos_up} show that TAPTRv2 obtains the best performance in all of the datasets' comprehensive metric AJ. Meanwhile, the consistent improvement of OA on all datasets further verifies the importance of our designs in keeping content feature uncontaminated for more accurate visibility classification. 
Note that, although the concurrent BootsTAP~\cite{doersch2024bootstap} obtains remarkable performance on Kinetics, it introduces extra 15M real world video clips for training. Moreover, we still outperform BootsTAP by about 2.1 AJ on the DAVIS dataset.

\vspace{-3mm}
\subsection{Ablation Studies and Analysis}
\vspace{-2mm}
We conduct ablation studies for each key design in our main contribution to gain a deeper understanding of what specifically contributes to performance improvement. We also perform ablation on the number of decoder layers.

\noindent \textbf{Ablation On The Introduction of Key-Aware Attention.} 
We take the type of attention mechanism in cross-attention as the only variable. The results in Table~\ref{tab:abl_attn_pos_up} show that (Row 2 vs. Row 1), the introduction of the key-aware deformable attention brings 0.7 AJ improvement, which is significant. The improvement indicates that the robust attention weights obtained through dot-production helps cross-attention obtain better query results from image feature maps, thereby improving the quality of point queries' content features.

\noindent \textbf{Ablation On The Position Update.} To verify the effectiveness of enabling the key-aware attention weights to function in the positional part of point queries, we conduct ablation studies as shown in Table~\ref{tab:abl_attn_pos_up}. The results (Row 3 vs. Row 2) show that using the attention weights for updating both the content and positional parts leads to a significant improvement (1.0 AJ). 
% This improvement indicates that make attention weights to function in the position part is reasonable. 
This improvement verifies that the local correlation information helps position estimation greatly, and our proposed attention-based position update is an effective operation to utilize correlation information.

\noindent \textbf{Ablation On The Weight Disentangling.}
As shown in Table~\ref{tab:abl_attn_pos_up}, decoupling the attention weights used for updating the content feature and position of a point query through an MLP enhances performance (0.9 AJ) (Row 4 vs. Row 3). This results verify that the attention weights required for the content and position parts may have different distributions, and simply mixing them confuses the network and may lead to sub-optimal results.

\noindent \textbf{Ablation On The Additional Supervision.}
To guarantee the attention-based position update in cross attention is always beneficial, it is important to supervise the updated positions in each decoder layer additionally. 
Results in Table~\ref{tab:abl_attn_pos_up} show that this extra supervision leads to a significant improvement (0.9 AJ) (Row 5 vs. Row 4), verifying its importance.

\begin{table*}[t]
\begin{center}
\resizebox{0.8\linewidth}{!}{ %
\begin{tabular}{c|c|c|c|c|ccc}
\toprule
Row & Key-Aware & Pos. Update. & Disentangle A. W.   & Supervision          & AJ   & $<\delta^{x}_{avg}$ & OA \\
\midrule
1 & \ding{55} &\ding{55}      & \ding{55}    & \ding{55}                  & 60.0 & 73.1                & 88.6 \\
2 & \checkmark &\ding{55}      & \ding{55}  & \ding{55}                    & 60.7 & 73.9                & 89.9 \\
3 & \checkmark &\checkmark      & \ding{55}   & \ding{55}                  & 61.7 & 74.8                & 90.4 \\
4 & \checkmark &\checkmark      & \checkmark   & \ding{55}                 & 62.6 & 75.5                & 91.0 \\
5 & \checkmark &\checkmark      & \checkmark   & \checkmark                & \textbf{63.5} & \textbf{75.9}               & \textbf{91.4} \\
\bottomrule
\end{tabular}
}
\caption{Ablation on each key design of the attention-based position updating. ``Pos.'' is short for ``Position'', and ``A. W.'' for ``Attention Weights.  }
\vspace{-7mm}
\label{tab:abl_attn_pos_up}
\end{center}
\end{table*}

\vspace{-2mm}
\begin{wraptable}{r}{7cm}
	\centering
    \resizebox{0.8\linewidth}{!}{
	\begin{tabular}{c|ccc}
    \toprule
    \#Decoder Layers               & AJ  & $<\delta^{x}_{avg}$ & OA \\
    \midrule
    2                             & 56.9 & 70.7 & 88.2 \\ 
    3                             & 60.3 & 74.0 & 89.8 \\
    4                             & 62.3 & 75.2 & 90.3 \\
    5                             & \textbf{63.5} & \textbf{75.9} &\textbf{91.4}  \\
    6                             & 62.7 & 75.7 &90.7  \\
    \bottomrule
    \end{tabular}
    }
    \caption{Ablation on the number of decoder layers.}
\label{tab:abl_deocer_layer}
\end{wraptable}

\noindent \textbf{Ablation On The Number of Decoder Layers.}
Since our improvements over TAPTR mainly focus on the decoder, we conduct ablation studies on the number of decoder layers to verify whether TAPTRv2 still satisfies the conclusion drawn from TAPTR. The results shown in Table~\ref{tab:abl_deocer_layer} indicate that, the performance of TAPTRv2 also improves with increased number of decoder layers, but reaches optimal performance with five decoder layers. This may be because, with the help of the additional position update, fewer decoder layers are needed for an optimal position update result.

\vspace{-3mm}

\section{Visualization}

\vspace{-3mm}

% \noindent\textbf{Comparison.} 
% As shown in Fig.~\ref{fig.compare_vis}, 
% when the dog turns around, CoTracker shows a significant drifting, where the tracking result shifts from the right side to the top of the dog. By contrast, {\methodname} tracks stably even when the tracking target is occluded.
% For more comparisons, fancy visualizations, and corresponding videos, please refer to our supplementary material.

\begin{figure*}[h]
    \vspace{-0.1mm}
    \centering
        \includegraphics[width=0.8\linewidth]{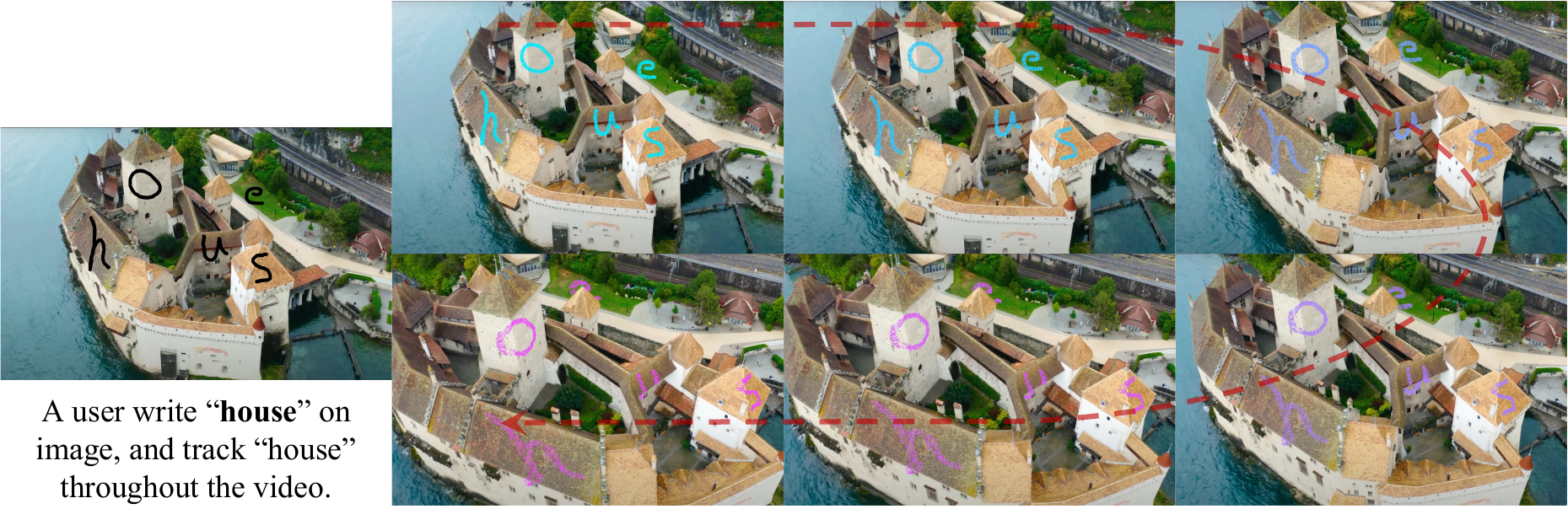}
    \caption{
    Visualization of the tracking results of TAPTRv2 in the wild. A user writes ``house'' on one frame and requires TAPTRv2 to track the points in the writing area. Best view in electronic version.
    }
\label{fig.in_the_wild_vis}
\end{figure*}

% \noindent\textbf{Stable Tracking Results In The Wild.} 
As shown in Fig.~\ref{fig.in_the_wild_vis}, {\methodname} shows its stability in point tracking and potential application in 3D reconstruction as well as video editing. For more visualizations and corresponding videos please refer to our supplementary materials.

% \vspace{-2mm}
% \begin{figure*}[h]
%     \vspace{-0.1mm}
%     \centering
%         \includegraphics[width=1\linewidth]{imgs/compare_vis.pdf}
    
%     \vspace{-2mm}
%     \caption{
%     Visualization of tracking results of TAPTR and TAPTRv2.
%     Solid and hollow red circles represent visible and invisible, respectively. We manually supplement the ground truth location of invisible points with blue crosses for better comparison. Best view in electronic version.
%     }
% \label{fig.compare_vis}
% \end{figure*}

\vspace{-3mm}

\vspace{-2mm}

\vspace{-3mm}

\section{Conclusion and Limitation}
\label{sec:conclusion}
\vspace{-3mm}
In this paper, 
we have presented {\methodname}, a new approach for solving the TAP task. TAPTRv2 improves TAPTR by developing a novel attention-based position update operation to address the query content feature contamination problem caused by the inappropriate integration of cost-volume in TAPTR. This operation is based on the observation that local attention is essentially the same as cost-volume, both of which are computed by dot-production between a query and its surrounding features. With this new operation, {\methodname} not only removes extra burden of cost-volume computation, but also leads to a substantial performance improvement. Compared with TAPTR, {\methodname} further simplifies the Transformer-based TAP framework, which we hope will help the TAP community scale up the training process and accelerate the development of more practical TAP algorithms. 
\\\noindent\textbf{Limitation and Future work.} For self-attention in our decoder, we currently use vanilla attention, which suffers from a computational cost quadratic to the number of queries. However, there have been many studies to reduce this cost to near linear. 
We will devote future research to solving it for a larger impact on dense point tracking. Additionally, TAPTRv2 aligns the frameworks of point tracking and object detection, which will facilitate the integration of multiple tasks. This will also be a topic we aim to address in the future.
\newpage

\appendix

\section{Some more discussions.}
\subsection{Different attention weight distribution requirements.}
\begin{figure}[h]
    \centering
    \includegraphics[width=0.4\linewidth]{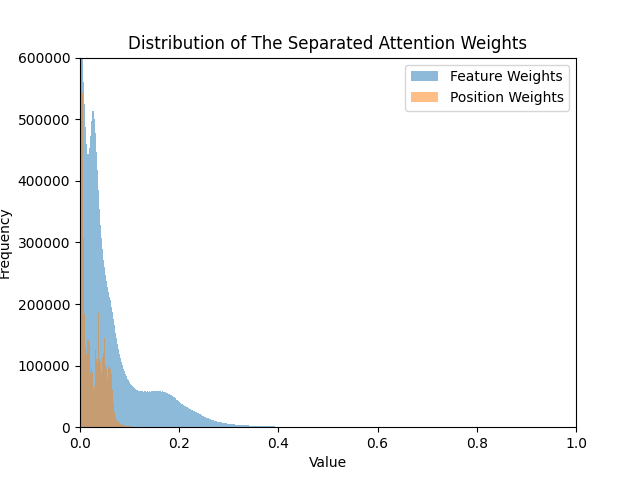}
    \caption{The visualization of the attention weight distributions for feature and position updating in our cross attention.  }
    \label{fig.attn_weight_distributions}
\end{figure}

We measured the distributions of the attention weights for content and position update in our cross attention, as visualized in Fig. \ref{fig.attn_weight_distributions}, the distributions of these two groups of attention weights show a significant difference, indicating that the attention weights required by content and position update are different. This can also verify the importance of our weight disentangle design in APU.

\vspace{-4mm}

\vspace{-5mm}

\subsection{Removing of cost-volume makes framework lightweight.}

TAPTRv2 exhibits a faster speed and lower resource requirements compared to TAPTR. More importantly, it's a common case that in the downstream tasks, we need to track all pixels in a region (e.g. tracking a text written on the back of a horse) rather than just a few scattered points. In this case, the number of points to be tracked will reach to tens of thousands. However, since the computation of the cost-volume and also the cost-volume aggregation operation in TAPTR increases sharply with the number of tracking points, with the number of tracking points increased, the advantage of TAPTRv2 will become more and more pronounced. As shown in the following second table, when the number of tracking points reaches 5000 (which is only 1.9\% of the pixels in a 512x512 image), the advantage of TAPTRv2 in speed and resource consumption becomes much more significant (about 24\% faster and 20\% fewer computational resource requirements).
We evaluate TAPTR and TAPTRv2 on A100 GPU (80G), and the computational efficiency (GFLOPS) is calculated following \href{https://github.com/facebookresearch/detectron2/blob/master/detectron2/utils/analysis.py}{detectron2}.

\vspace{-3mm}
\begin{table}[h]
    \centering
    \resizebox{0.7\linewidth}{!}{
    \begin{tabular}{c|ccc}
        % First table
        \begin{tabular}{c|ccc}
        \toprule
        800 Points & FPS & GFLOPS & \#Param  \\ 
        \midrule
        TAPTR      & 65.9 & 147.2 & 39.2M  \\ 
        TAPTRv2    & 69.1 & 143.4 & 38.2M \\
        \bottomrule
        \end{tabular}
        &
        % Second table
        \begin{tabular}{c|ccc}
        \toprule
        5000 Points & FPS & GFLOPS & \#Param  \\ 
        \midrule
        TAPTR      & 11.8 & 426.8 & 39.2M  \\ 
        TAPTRv2    & 14.6 & 354.2 & 38.2M \\
        \bottomrule
        \end{tabular}
    \end{tabular}
    }
    \caption{Comparison of speed, computational efficiency, and resource requirements between TAPTR and TAPTRv2.}
\end{table}

\vspace{-3mm}

\section{More Visualizations}

\subsection{Application of {\methodname} in Video Editing}
Here we show the results of the video editing using {\methodname}. After the users plot on one frame to specify the region to be edited, we sample points in the editing area and track these points across the whole video. For more details please refer to the videos in our supplementary material. Corresponding video names are provided in Sec.~\ref{sec.video_correspondences}.

\begin{figure}[h]
    \centering
    \includegraphics[width=1\linewidth]{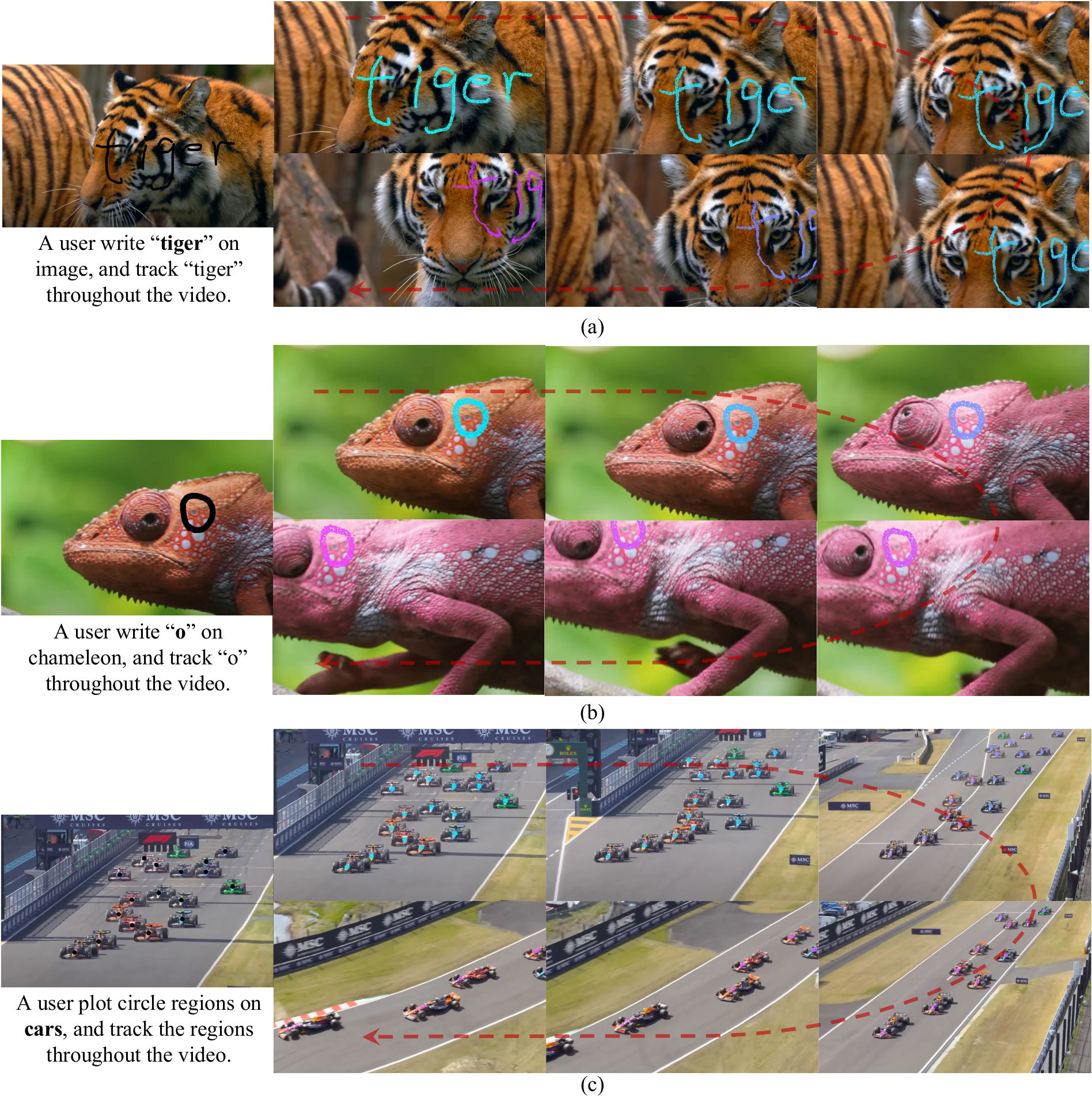}
    \caption{Apply {\methodname} in Video Editing. The color of the editing area changes over time.  }
    \label{fig.video_edit}
\end{figure}

Fig.~\ref{fig.video_edit} (a) not only shows the ability of video editing but also the potential of {\methodname} in applying in 3D reconstruction.

Fig.~\ref{fig.video_edit} (b) shows that {\methodname} can handle the color change during the tracking. More importantly, although the editing area is cluttered in the middle of the video {\methodname} can robustly continue tracking the editing area when it reappears again.

Fig.~\ref{fig.video_edit} (c) shows that {\methodname} has the ability to handle the changing of scale.

\subsection{Application of {\methodname} in Trajectory Estimation}
\begin{figure}[h]
    \centering
    \includegraphics[width=1\linewidth]{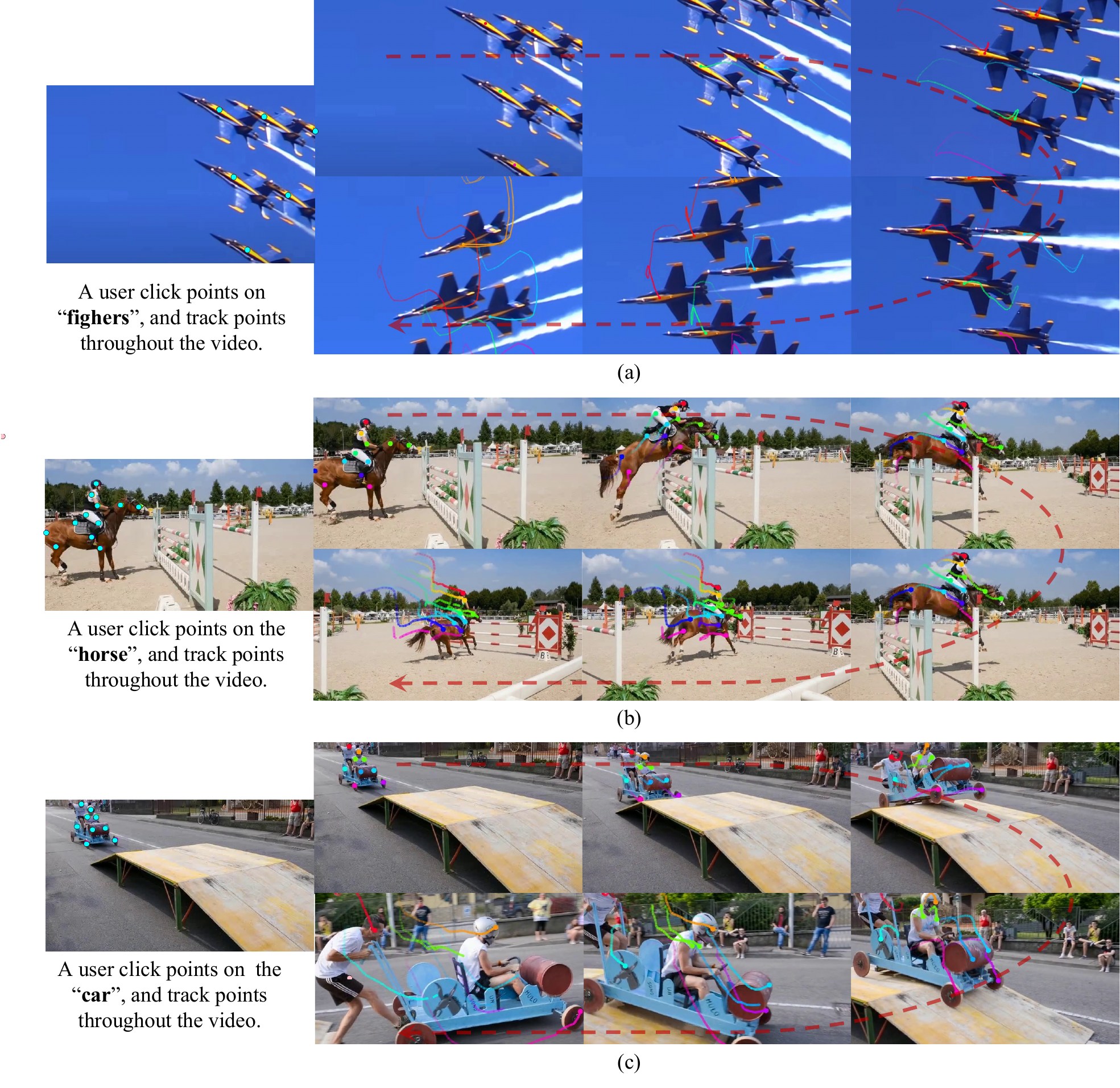}
    \caption{Apply {\methodname} in Trajectory Estimation.}
    \label{fig.trajectory}
\end{figure}

In Fig.~\ref{fig.trajectory} we show the results of the trajectory estimation using {\methodname}. After the users click on one frame to specify the points to be tracked, {\methodname} will keep tracking these points across the whole video to construct their trajectory.

% \section{Correspondences Between Visualizations and Videos}
% \label{sec.video_correspondences} 

% \noindent Fig.~\ref{fig.video_edit} (a) --> VideoEdit\_house.mp4 \\
% \noindent Fig.~\ref{fig.video_edit} (b) --> VideoEdit\_chameleon.mp4 \\
% \noindent Fig.~\ref{fig.video_edit} (c) --> VideoEdit\_F1.mp4 \\
% \noindent Fig.~\ref{fig.trajectory} (a) --> Trajectory\_fighter.mp4 \\
% \noindent Fig.~\ref{fig.trajectory} (b) --> Trajectory\_horsejump.mp4 \\
% \noindent Fig.~\ref{fig.trajectory} (c) --> Trajectory\_soapbox.mp4 \\

\clearpage

\bibliographystyle{splncs04}
\bibliography{mainbib}
\end{document}